\documentclass[letterpaper, 10 pt, conference]{ieeeconf}  % Comment this line out if you need a4paper
\IEEEoverridecommandlockouts                              % This command is only needed if 
\usepackage{graphics} % for pdf, bitmapped graphics files
\usepackage{epsfig} % for postscript graphics files
\usepackage{times} % assumes new font selection scheme installed
\usepackage{amsmath} % assumes amsmath package installed
\usepackage{amssymb}  % assumes amsmath package installed
\usepackage{multirow}
\usepackage{pdfpages}
\usepackage{subfigure}
\usepackage{bm}
\usepackage{colortbl}
\usepackage{gensymb}
\usepackage{url}
\usepackage{comment}
\usepackage{booktabs}
\usepackage{color}
\usepackage{cite}
\usepackage{multido}%
\usepackage[colorlinks,
            linkcolor=blue,
            anchorcolor=red,
            citecolor=green
            ]{hyperref}

\makeatletter
\newcommand{\ssymbol}[1]{^{\@fnsymbol{#1}}}
\makeatother

\begin{document}

\title{\LARGE \bf
PandaSet: Advanced Sensor Suite Dataset for Autonomous Driving
}

\author{Pengchuan Xiao$^{1}$, Zhenlei Shao$^{1,\ast}$, Steven Hao$^{2}$, Zishuo Zhang$^{3}$, Xiaolin Chai$^{1}$, Judy Jiao$^{1}$, 
\\Zesong Li$^{1}$, Jian Wu$^{1}$, Kai Sun$^{1}$, Kun Jiang$^{4}$, Yunlong Wang$^{4}$ and Diange Yang$^{4,\ast}$ % <-this % stops a space
\thanks{$^{1}$Pengchuan Xiao, Zhenlei Shao, Xiaolin Chai, Judy Jiao, Zesong Li, Jian Wu, and Kai Sun are with Hesai Technology Co., Ltd., China. 
        {\tt\small\{xiaopengchuan, shaozhenlei, chaixiaolin, judy.jiao, lizesong, wujian, s\}@hesaitech.com}}
\thanks{$^{2}$Steven Hao is with Scale AI, Inc., USA. 
        {\tt\small{steven}@scale.com}}%
\thanks{$^{3}$Zishuo Zhang is with Princeton University, USA.
        {\tt\small{zishuoz}@princeton.edu}}%
\thanks{$^{4}$Kun Jiang, Yunlong Wang, and Diange Yang are with State Key Laboratory of Automotive Safety and Energy, Center for Intelligent Connected Vehicles and Transportation, School of Vehicle and Mobility, Tsinghua University, China.
        {\tt\small{jiangkun}@tsinghua.edu.cn, { yl-wang19}@mails.tsinghua.edu.cn, {ydg}@tsinghua.edu.cn}}%
\thanks{*Corresponding to {\tt\small{Zhenlei Shao}@hesaitech.com, {ydg}@tsinghua.edu.cn}}% <-this % stops a space
}

\maketitle
%\twocolumn[{
%	\renewcommand\twocolumn[1][]{#1}
%	\maketitle
%	\begin{center}
%		\vspace{-0.6cm}
%		\includegraphics[width=\textwidth]{image/headimage.pdf}
%       \vspace{-0.65cm}
%		%\captionof{figure}
%		{
%			
%               }
%		\label{fig:teaser}
%	\end{center}
%	\vspace{0.4cm}
%}]
%\maketitle
\thispagestyle{empty}
\pagestyle{empty}

%%%%%%%%%%%%%%%%%%%%%%%%%%%%%%%%%%%%%%%%%%%%%%%%%%%%%%%%%%%%%%%%%%%%%%%%%%%%%%%%
\begin{abstract}

The accelerating development of autonomous driving technology has placed greater demands on obtaining large amounts of high-quality data. Representative, labeled, real world data serves as the fuel for training deep learning networks, critical for improving self-driving perception algorithms. In this paper, we introduce PandaSet, the first dataset produced by a complete, high-precision autonomous vehicle sensor kit with a no-cost commercial license. The dataset was collected using one $360\degree$ mechanical spinning LiDAR, one forward-facing, long-range LiDAR, and $6$ cameras. The dataset contains more than $100$ scenes, each of which is $8$ seconds long, and provides $28$ types of labels for object classification and $37$ types of labels for semantic segmentation. We provide baselines for LiDAR-only 3D object detection, LiDAR-camera fusion 3D object detection and LiDAR point cloud segmentation. For more details about PandaSet and the development kit, see https://scale.com/open-datasets/pandaset.

\end{abstract}

%%%%%%%%%%%%%%%%%%%%%%%%%%%%%%%%%%%%%%%%%%%%%%%%%%%%%%%%%%%%%%%%%%%%%%%%%%%%%%%%
\section{INTRODUCTION}
\label{sec:INTRODUCTION}
Autonomous driving has attracted widespread attention in recent years with its potential to fundamentally disrupt the transportation and mobility landscape. A key component of the autonomous driving technology stack is 3D perception technology. Based on the current state of machine learning, 3D perception relies on large amounts of high-quality, real-world annotated data\cite{Caesar_2020_CVPR}. The data needs to satisfy two requirements. First, the sensors used for data collection demand sufficiently high precision. If sensor-collected data fails to achieve high precision (due to imprecise LiDAR range measurements, pixelated camera images, etc.), the performance of the corresponding back-end algorithm developed using this data will be limited\cite{kim2018robust}. Second, the ground truth labels need to be sufficiently accurate and complete. For current data-driven machine learning methods, incorrect and incomplete labeling might even deteriorate the performance of the model. For example, in object detection, a large number of polluted labels have been shown to hurt accuracy\cite{feng2020labels}. The demand for high-quality data also encompasses requirements on the diversity and complexity of the captured scenes. Autonomous driving is currently concentrated in limited, geofenced areas. But complex environments, whether due to different lighting conditions, changing traffic flow, hazardous road conditions, complex vegetation, unexpected human movements or positions, or unfamiliar objects are all potential problems in real-world driving scenarios\cite{guo2019safe}. Datasets that capture richer and more diverse scenes, or provide different levels of annotation, can help improve the robustness of autonomous vehicles\cite{pham20203d}. However, due to the impact of COVID-19, a large number of autonomous driving companies had to suspend their road testing in 2020, which led to a significant reduction of road test data. To help fill this gap, we launched PandaSet: an open-source dataset for training autonomous driving machine learning models. We hope that PandaSet will serve as a valuable resource to promote and advance research and development in autonomous driving and machine learning.

The main contributions of this paper are listed as follows:

\begin{itemize}
        \item We present a multimodal dataset named PandaSet, which provides a complete kit of high-precision sensors covering a $360\degree$ field of view. It is the world's first open-source dataset to feature both mechanical spinning and forward-facing LiDARs and to be licensed for free without major restrictions on its research or commercial use.
        \item PandaSet features $28$ different annotation classes for each scene as well as $37$ semantic segmentation labels for most scenes. All of the annotations are labeled under multi-sensor fusion to ensure that each ground truth label is sufficiently accurate and precise.
        \item PandaSet includes data from complex metropolitan driving environments: traffic and pedestrians, construction zones, hills, and varied lighting conditions throughout the day and at night. It covers challenging driving conditions for full level 4 and 5 driving autonomy. There is a high density of useful information, with many more objects in each frame than in other datasets.
        \item Based on PandaSet, we provide baselines for LiDAR-only 3D object detection, LiDAR-camera fusion 3D object detection, and LiDAR point cloud segmentation, and a corresponding devkit for researchers to use the dataset directly.
\end{itemize}

\begin{table*}[t]
        \scriptsize
        \caption{Current autonomous driving datasets.}
        % \vspace{-15pt}
        \renewcommand\arraystretch{1.5}
        {\centering
        % \begin{center}
        % \vspace{-15pt}
        \label{tab:Current autonomous driving datasets}
        \begin{tabular}{|p{1.4cm}<{\centering}|p{0.4cm}<{\centering}p{0.6cm}<{\centering}p{1cm}<{\centering}p{0.3cm}<{\centering}p{1.0cm}<{\centering}p{0.3cm}<{\centering}p{0.5cm}<{\centering}p{0.6cm}<{\centering}p{1.2cm}<{\centering}p{0.6cm}<{\centering}p{0.6cm}<{\centering}p{1.1cm}<{\centering}p{1.8cm}<{\centering}|}
        %\multicolumn{14}{c}{\footnotesize TABLE I~Current autonomous driving datasets} \\
        % \multicolumn{14}{l}{\textbf{Notes}:()$\ssymbol{2}$: (3D object detection)/(semantic segmentation) ()$\ssymbol{3}$: (semantic 2D instances segmentation) / (per-pixel semantic labeling); MS: mechanical spinning}\\
        \multicolumn{14}{l}{\textbf{Notes}:()$\ssymbol{2}$: (3D Object detection)/(Semantic segmentation); ()$\ssymbol{3}$: (Semantic 2D instances segmentation)/(Per-pixel semantic labeling); ()$\ssymbol{9}$: Only ground;}\\
        \multicolumn{14}{l}{()$\ssymbol{8}$: (3D Object detection)/(Point cloud semantic segmentation); MS: Mechanical spinning; FF: Forward-facing; C: Cameras; L: LiDARs; R: Radars; G\&I: GNSS\&IMU.}\\
        % \multicolumn{14}{l}{FF: forward-facing ; ()$\ssymbol{8}$: (3D object detection)/(point cloud semantic segmentation); ()$\ssymbol{9}$: Only ground}\\
        \hline
        \multirow{2}{*}[0.5ex]{\textbf{Dataset}} &\multirow{2}{*}[0.5ex]{\textbf{Year}} & \multirow{2}{*}[0.5ex]{\textbf{Scenes}} &\textbf{Ann. Frames}         & \multirow{2}{*}[0.5ex]{\textbf{C}}             & \multirow{2}{*}[0.5ex]{\textbf{L}}                   & \multirow{2}{*}[0.5ex]{\textbf{R}}                 & \multirow{2}{*}[0.5ex]{\textbf{G\&I}} & \textbf{3D Boxes} & \multirow{2}{*}[0.5ex]{\textbf{Classes}}           & \textbf{LiDAR Seg.}                  & \multirow{2}{*}[0.5ex]{\textbf{Night}}     & \multirow{2}{*}[0.5ex]{\textbf{Locations}}   & \textbf{No-Cost Comm- ercial License}   \\
        \hline
        \hline
        % \multirow{1}{*}{KITTI}                  & \multirow{1}{*}{2012}                         & \multirow{1}{*}{22}                  & \multirow{1}{*}{7481}                                     & FF                  & \multirow{1}{*}{1x MS}                    & \multirow{1}{*}{$\times$}                     & \multirow{1}{*}{\checkmark}      & \multirow{1}{*}{\checkmark}      & \multirow{1}{*}{8}                 & \multirow{1}{*}{$\times$}                 & \multirow{1}{*}{$\times$}        & \multirow{1}{*}{Karlsruhe}                                    & \multirow{1}{*}{$\times$}                                                       \\ \hline
        KITTI                  & 2012                         & $22$                  & $7481$                                     & FF                  & $1\times$ MS                    & $\times$                     & \checkmark      & \checkmark      & $8$                 & $\times$                 & $\times$        & Karlsruhe                                    & $\times$                                                       \\ \hline
        ApolloScape            & 2018                         & -                   & $144k$                                     & FF                  & $2\times$ MS                    & $\times$                     & \checkmark      & \checkmark      & $8/35\ssymbol{3}$           & \checkmark                & \checkmark       & $4\times$ China                                     & $\times$                                                       \\ \hline
        % \multirow{1}{*}{ApolloScape}            & \multirow{1}{*}{2018}                         & \multirow{1}{*}{-}                   & \multirow{1}{*}{144k}                                     & FF                  & \multirow{1}{*}{2x MS}                    & \multirow{1}{*}{$\times$}                     & \multirow{1}{*}{\checkmark}      & \multirow{1}{*}{\checkmark}      & \multirow{1}{*}{8/35$\ssymbol{3}$}            & \multirow{1}{*}{\checkmark}                & \multirow{1}{*}{\checkmark}       & \multirow{1}{*}{4x China}                                     & \multirow{1}{*}{$\times$}                                                       \\ \hline
        \multirow{2}{*}[0.25ex]{nuScenes}               & \multirow{2}{*}[0.25ex]{2019}                         & \multirow{2}{*}[0.25ex]{$\mathbf{1k}$}                  & \multirow{2}{*}[0.25ex]{$40k$}                                      & \multirow{2}{*}[0.25ex]{\textbf{360$^\circ$}}         & \multirow{2}{*}[0.25ex]{$1\times$ MS}                    & \multirow{2}{*}[0.25ex]{\checkmark}                    & \multirow{2}{*}[0.25ex]{\checkmark}      & \multirow{2}{*}[0.25ex]{\checkmark}      & \multirow{2}{*}[0.25ex]{$23$}                & \multirow{2}{*}[0.25ex]{\checkmark}                & \multirow{2}{*}[0.25ex]{\checkmark}       & Boston  Singapore                                             & \multirow{2}{*}[0.25ex]{$\times$}                                                   \\ \hline
        %Argoverse              & 2019                         & 113                 & 22k                                      & \textbf{360$^\circ$}         & 2x MS                    & $\times$                     & \checkmark      & \checkmark      & 15                & \checkmark$\ssymbol{9}$             & \checkmark       & Pittsbirgh  Miami                                             & $\times$                                                  \\ \hline
        \multirow{2}{*}[0.25ex]{Argoverse}              & \multirow{2}{*}[0.25ex]{2019}                         & \multirow{2}{*}[0.25ex]{$113$}                 & \multirow{2}{*}[0.25ex]{$22k$}                                      & \multirow{2}{*}[0.25ex]{\textbf{360$^\circ$}}         & \multirow{2}{*}[0.25ex]{$2\times$ MS}                    & \multirow{2}{*}[0.25ex]{$\times$}                     & \multirow{2}{*}[0.25ex]{\checkmark}      & \multirow{2}{*}[0.25ex]{\checkmark}      & \multirow{2}{*}[0.25ex]{$15$}                & \multirow{2}{*}[0.25ex]{\checkmark$\ssymbol{9}$}             & \multirow{2}{*}[0.25ex]{\checkmark}       & Pittsburgh  Miami                                             & \multirow{2}{*}[0.25ex]{$\times$}                                                  \\ \hline
        % \multirow{1}{*}{Lyft L5}                & \multirow{1}{*}{2019}                         & \multirow{1}{*}{366}                 & \multirow{1}{*}{46k}                                      & \multirow{1}{*}{\textbf{360$^\circ$}}         & \multirow{1}{*}{3x MS}                    & \multirow{1}{*}{$\times$}                     & \multirow{1}{*}{\checkmark}      & \multirow{1}{*}{\checkmark}      & \multirow{1}{*}{9}                 & \multirow{1}{*}{$\times$}                 & \multirow{1}{*}{$\times$}        & \multirow{1}{*}{Palo Alto}                                    & $\times$                                                                        \\ \hline
        Lyft L5                & 2019                         & $366$                 & $46k$                                      & \textbf{360$^\circ$}         & $3\times$ MS                    & $\times$                     & \checkmark      & \checkmark      & $9$                 & $\times$                 & $\times$        & Palo Alto                                    & $\times$                                                                        \\ \hline
        Waymo Open             & 2019                       & $\mathbf{1k}$                  & $\mathbf{200k}$                                     & \textbf{360$^\circ$}         & $\mathbf{5\times}$ \textbf{MS}                   & $\times$                     & \checkmark      & \checkmark      & $4$                 & $\times$                 & \checkmark       & $3\times$ USA                                                              & $\times$                                                       \\ \hline
        % \multirow{1}{*}{A$\ast$3D}              & \multirow{1}{*}{2019}                         & \multirow{1}{*}{-}                   & \multirow{1}{*}{39k}                                      & FF                  & \multirow{1}{*}{1x MS}                    & \multirow{1}{*}{$\times$}                     & \multirow{1}{*}{\checkmark}      & \multirow{1}{*}{\checkmark}      & \multirow{1}{*}{7}                 & \multirow{1}{*}{$\times$}                 & \multirow{1}{*}{\checkmark}       & \multirow{1}{*}{Singapore}                                    & \multirow{1}{*}{$\times$}                                                       \\ \hline
        A$\ast$3D              & 2019                         & -                   & $39k$                                      & FF                  & $1\times$ MS                    & $\times$                     & \checkmark      & \checkmark      & $7$                 & $\times$                 & \checkmark       & Singapore                                    & $\times$                                                       \\ \hline
        % \multirow{2}{*}{\textbf{PandaSet}}               & \multirow{2}{*}{2020}                         & \multirow{2}{*}{103/76$\ssymbol{2}$}                         & \multirow{2}{*}{8240/6080$\ssymbol{2}$}                                            & \multirow{2}{*}{\textbf{360$^\circ$}}         & \textbf{1x MS 1x FF}                               & \multirow{2}{*}{\textbf{$\times$}}                     & \multirow{2}{*}{\textbf{\checkmark}}      & \multirow{2}{*}{\textbf{\checkmark}}      & \multirow{2}{*}{\textbf{28/37$\ssymbol{8}$}}                          & \multirow{2}{*}{\textbf{$\times$}}                 & \multirow{2}{*}{\textbf{\checkmark}}       & \multirow{2}{*}{San Francisco}  & \multirow{2}{*}{\textbf{\checkmark}} \\ \hline
        \multirow{2}{*}[0.25ex]{\textbf{PandaSet}}               & \multirow{2}{*}[0.25ex]{2020}                         & $103$/ $76\ssymbol{2}$                         & $8240$/ $6080\ssymbol{2}$                                            & \multirow{2}{*}[0.25ex]{\textbf{360$^\circ$}}         & $\mathbf{1\times}$ \textbf{MS} $\mathbf{1\times}$ \textbf{FF}                               & \multirow{2}{*}[0.25ex]{\textbf{$\times$}}                     & \multirow{2}{*}[0.25ex]{\textbf{\checkmark}}      & \multirow{2}{*}[0.25ex]{\textbf{\checkmark}}      & \multirow{2}{*}[0.25ex]{$\mathbf{28}$/$\mathbf{37\ssymbol{8}}$}                          & \multirow{2}{*}[0.25ex]{\textbf{$\times$}}                 & \multirow{2}{*}[0.25ex]{\textbf{\checkmark}}       & San Francisco  & \multirow{2}{*}[0.25ex]{\textbf{\checkmark}} \\ \hline
        % \multirow{1}{*}{Cirrus}                 & \multirow{1}{*}{2020}                         & \multirow{1}{*}{12}                  & \multirow{1}{*}{6285}                                     & FF                  & \multirow{1}{*}{2x FF}                    & \multirow{1}{*}{$\times$}                     & \multirow{1}{*}{\checkmark}      & \multirow{1}{*}{\checkmark}      & \multirow{1}{*}{8}                 & \multirow{1}{*}{$\times$}                 & \multirow{1}{*}{\checkmark}       & -                                                              & \multirow{1}{*}{\checkmark}                                                      \\ \hline
        Cirrus                 & 2020                         & $12$                  & $6285$                                     & FF                  & $2\times$ FF                    & $\times$                     & \checkmark      & \checkmark      & $8$                 & $\times$                 & \checkmark       & -                                                              & \checkmark                                                      \\ \hline
        \end{tabular} 
        % \end{center}
        \par}\medskip
    \vspace{-10pt} 
    \end{table*}

%%%%%%%%%%%%%%%%%%%%%%%%%%%%%%%%%%%%%%%%%%%%%%%%%%%%%%%%%%%%%%%%%%%%%%%%%%%%%%%%
\section{RELATED WORK}
\label{sec:RELATED WORK}

Over the past ten years, data-driven approaches to machine learning have become increasingly popular, leading to significant progress in the development of 3D perception technology\cite{chen2017multi,yan2018second,zhou2018voxelnet,lang2019pointpillars,chen2019fast,yang2019std,shi2019pointrcnn,yang20203dssd,shi2020pv} and implementation of corresponding autonomous driving applications. Numerous datasets\cite{geiger2012we,huang2018apolloscape,chang2019argoverse,lyft2019,sun2020scalability,pham20203d,wang2020cirrus} for autonomous driving have been released by organizations globally. We list current open-source datasets in Table \ref{tab:Current autonomous driving datasets}, considering only those that include both camera-collected and LiDAR-collected data, as well as 3D annotations. The pioneering KITTI dataset\cite{geiger2012we}, launched in 2012, is considered the first benchmark dataset collected for an autonomous driving platform. It features two stereo camera systems, a mechanical spinning LiDAR, and a GNSS/IMU device. However, in the KITTI dataset, an object is only annotated when it appears in the vehicle's front-view camera's field of view (FOV). It also only features daytime data. The 2018 ApolloScape dataset\cite{huang2018apolloscape} employs a sensor configuration similar to that of KITTI, but its LiDAR is placed slanted at the rear of the car. Such placement is primarily geared toward map data collection and provides static data on depth rather than complete point cloud information. The nuScenes\cite{Caesar_2020_CVPR}, Argoverse\cite{chang2019argoverse}, Lyft L5\cite{lyft2019}, Waymo Open\cite{sun2020scalability}, and A$\ast$3D datasets\cite{pham20203d} launched in 2019 expanded the availability and quality of open-source datasets. Among them, nuScenes, Argoverse and Lyft L5 added map data. nuScenes added additional radar data. However, Argoverse only provides point cloud semantic segmentation for one category. Lyft L5 does not include nighttime data. A$\ast$3D only provides front-facing camera data and does not provide annotations for point cloud semantic segmentation. nuScenes only provides point cloud within $70$ meters. Waymo's data collection vehicle was equipped with a 64-channel spinning LiDAR, but the point cloud provided is only within $75$ meters. The Cirrus dataset\cite{wang2020cirrus} launched in 2020 was equipped with a pair of long-range bi-pattern LiDARs with a 250-meter effective range in the front-facing direction. However, Cirrus did not employ any $360\degree$ FOV sensor, limiting its perception range to only the front-facing view. 
%\begin{figure}[]
%        \vspace{-10pt}
%        \begin{center}
%        \includegraphics[trim = 0mm 30mm 0mm 8mm, clip, width = 8.8cm]{image/2.eps}
%        \end{center}
%        \vspace{-20pt}
%        \caption{Pandar64 and PandarGT point cloud visualization example}
%        \label{fig:pc of LiDAR}
%        \vspace{-10pt}
%\end{figure}

%%%%%%%%%%%%%%%%%%%%%%%%%%%%%%%%%%%%%%%%%%%%%%%%%%%%%%%%%%%%%%%%%%%%%%%%%%%%%%%%

%\begin{figure}[t]
%        \vspace{-10pt}
%        \begin{center}
%          \subfigure[Sensors placement]{\label{fig:car}\includegraphics[scale=0.75]{image/car2.pdf}}
%          \subfigure[Sensors coverage]{\label{fig:coverage}\includegraphics[scale=0.7,angle=271]{image/view.pdf}}
%        \end{center}
%        \vspace{-10pt}
%        \caption{Sensors configuration}
%        \label{fig:sensor}
%\vspace{-20pt}
%\end{figure}

\begin{figure}[t]
        \centering
        \setlength{\abovecaptionskip}{-3pt}
        \subfigure[Sensor placement.]{
        \begin{minipage}[t]{1\linewidth}
        \centering
        \includegraphics[scale=0.2]{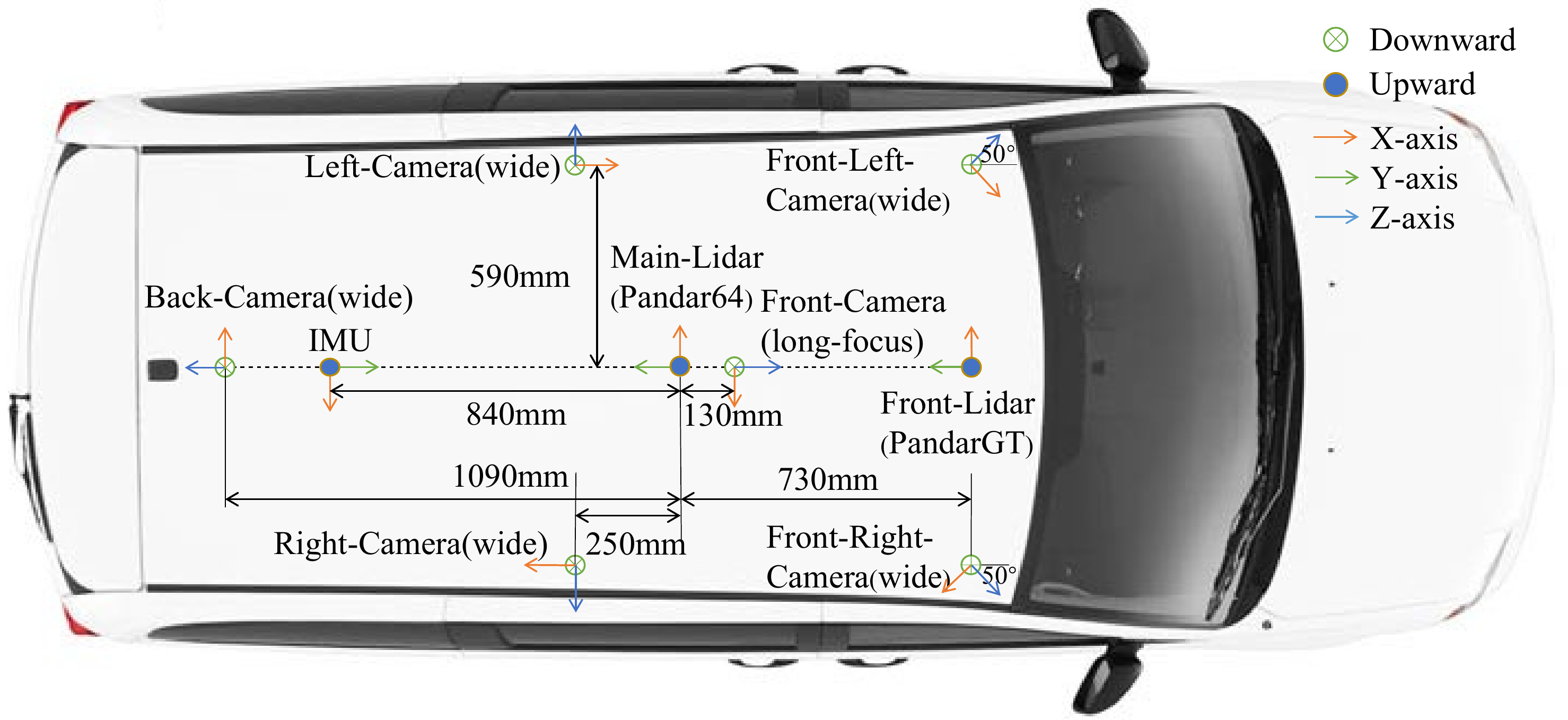}
        \end{minipage}
        }\\
        \subfigure[Sensor coverage.]{
        \begin{minipage}[t]{1\linewidth}
        \centering
        \includegraphics[scale=0.5]{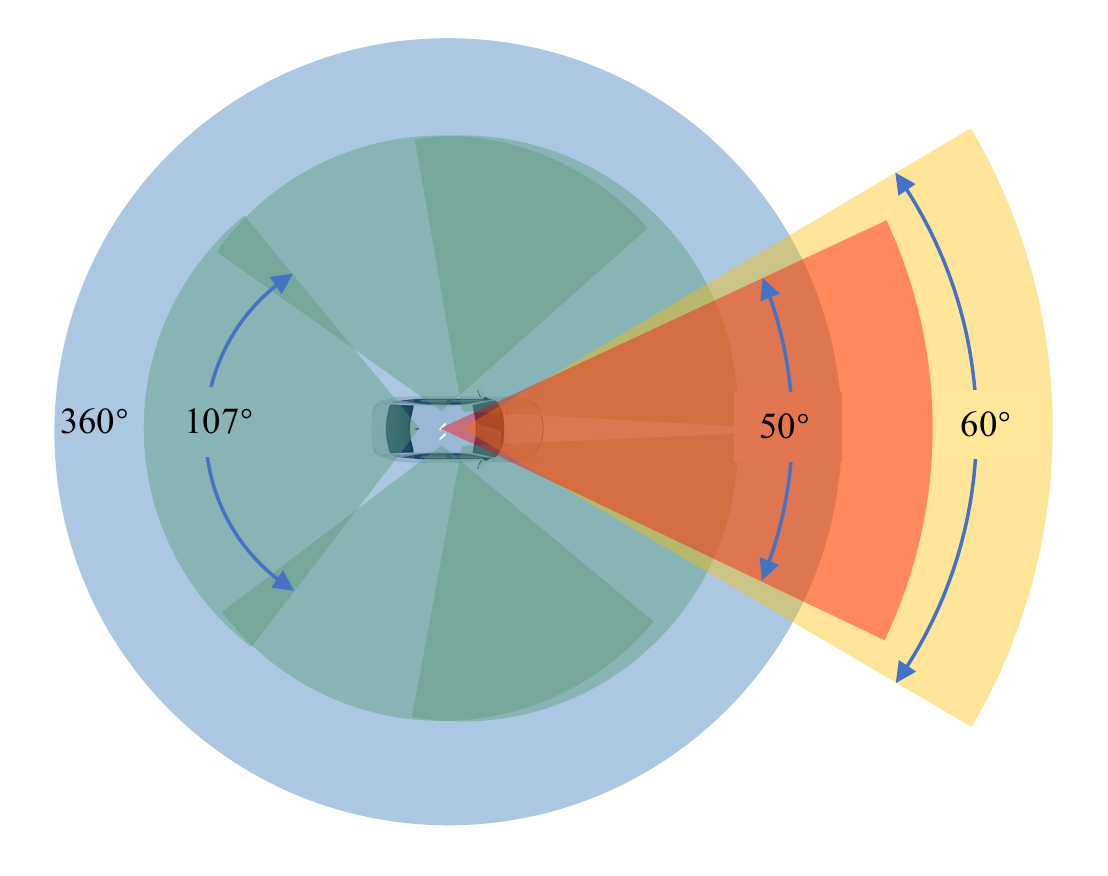}
        \end{minipage}
        }
        \centering
        \caption{Data collection sensor suite.} 
        %BC-W:Back-Camera(Wide); LC-W:Left-Camera(Wide); RC-W:Right-Camera(Wide); FC-LF:Front-Camera(Long Focus); FLC-W:Front-Left-Camera(Wide); FRC-W:Front-Right-Camera(Wide); ML:Main-Lidar; LL:Left-Lidar; RL:Right-Lidar; GT:PandarGT}
        \label{fig:Sensor Setup for dataset acquisition}
        \vspace{-16pt}
        \end{figure}

\begin{figure*}[h]
        \begin{center}
                \includegraphics[width=\textwidth]{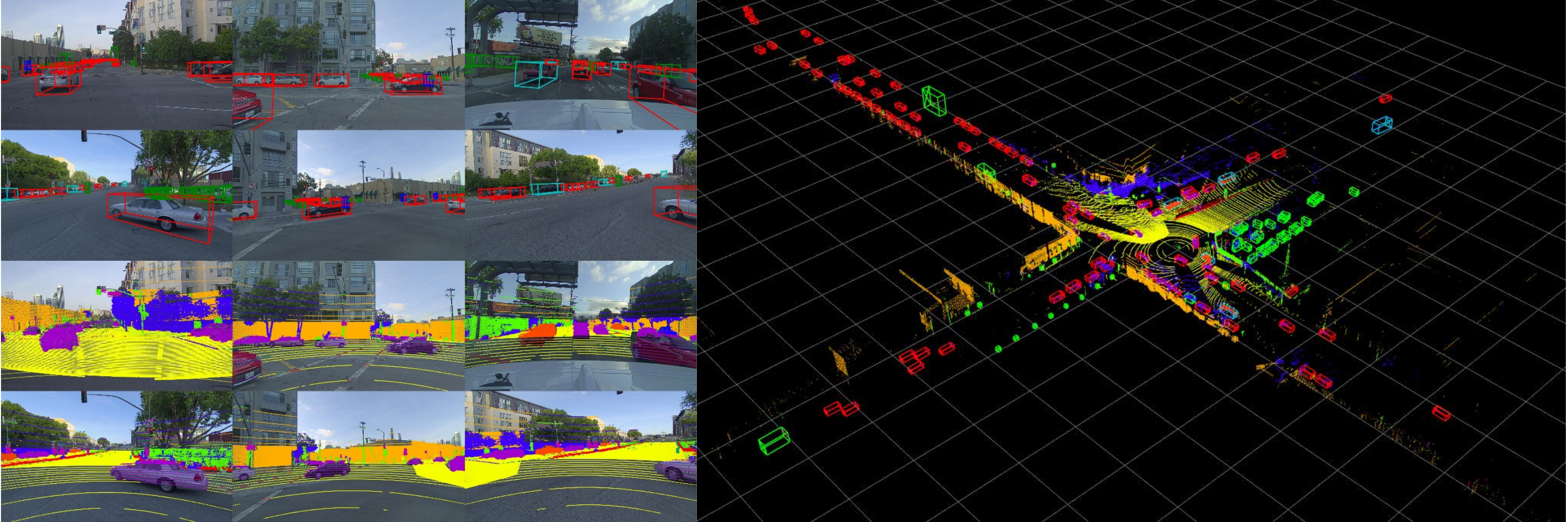}
        \end{center}
        \vspace{-10pt}
        \caption{A sample from the PandaSet dataset. Left: Camera images with projected 3D bounding box annotations (upper $6$ photos) and projected point cloud semantic segmentation annotations (lower $6$ photos), taken from a single $360\degree$ capture. Right: Combined point cloud of the forward-facing LiDAR and the mechanical spinning LiDAR, with annotations in 3D view. The length of each grid is equivalent to $20m$.}       
        \begin{center}
              %  \vspace{-5pt}
              %  \footnotesize{(grids in figure length 20m)}
        \end{center}
        \label{fig:Overview of PandaSet}
\vspace{-15pt}
\end{figure*}

\section{PANDASET DATASET}
\label{sec:PANDASET DATASET}

Here we introduce our methods for data collection, sensor calibration, and data annotation, then provide a brief analysis of our dataset. 

\begin{table*}[bhtp]
        \caption{sensor specifications}
        %\label{tab:The specifications of sensors}
        \renewcommand\arraystretch{1.5}
        \vspace{-10pt}
        \begin{center}
        \begin{tabular}{|l|p{12cm}|}
        %\multicolumn{2}{c}{\footnotesize TABLE \uppercase\expandafter{\romannumeral2~ The specifications of sensors}}   \\  
        \hline
        \multicolumn{1}{|c|}{\textbf{Sensor}}                    & \multicolumn{1}{c|}{\textbf{Details}}\\
        \hline
        \hline
        $1\times$ Mechanical spinning LiDAR                    & $360\degree$ horizontal FOV, $10Hz$, $64$ channels, $200m$ range @ $10$ \% reflectivity (Hesai Pandar64) \\
        \hline
        \multirow{2}{*}[0.25ex]{$1\times$ Forward-facing LiDAR}                         & $60\degree$ horizontal FOV, $10Hz$, equivalent to $150$ channels  @ $10Hz$, $300m$ range @ $10\%$ reflectivity (Hesai PandarGT) \\
        \hline
        %\multirow{3}{*}{5x Wide-angle cameras}          & 107$^\circ$ horizontal FOV, 10Hz, 1/2.7" CMOS sensor of 1920x1080 resolution, auto exposure, unpacked to YUV 4:4:4 format and compressed to JPEG (Leopard LI-USB30-AR023ZWDRB camera with M128B02820WM12R1 lens) \\
        \multirow{2}{*}[0.25ex]{$5\times$ Wide-angle cameras}          & $107\degree$ horizontal FOV, $10Hz$, $1/2.7"$ CMOS sensor of $1920\times1080$ resolution, auto exposure (Leopard LI-USB30-AR023ZWDRB camera with M128B02820WM12R1 lens) \\
        \hline
%       % \multirow{3}{*}{1x Forward-facing long-focus camera} & 50$^\circ$ horizontal FOV, 10Hz, 1/2.7" CMOS sensor of 1920x1080 resolution, auto exposure, unpacked to YUV 4:4:4 format and compressed to JPEG (Leopard LI-USB30-AR023ZWDRB camera with CS-6IR(5MP)-L lens) \\
        \multirow{2}{*}[0.25ex]{$1\times$ Forward-facing long-focus camera} & $50\degree$ horizontal FOV, $10Hz$, $1/2.7"$ CMOS sensor of $1920\times1080$ resolution, auto exposure (Leopard LI-USB30-AR023ZWDRB camera with CS-6IR(5MP)-L lens) \\
        % 1x Forward-facing long-focus camera & 50$^\circ$ horizontal FOV, 10Hz, 1/2.7" CMOS sensor of 1920x1080 resolution, auto exposure (Leopard LI-USB30-AR023ZWDRB camera with CS-6IR(5MP)-L lens) \\
        \hline 
        $1\times$ GNSS\&IMU                                     & (NovAtel PwrPak7) \\
        \hline
        \end{tabular}
        \end{center}
\vspace{-10pt}
\label{tab:The specifications of sensors}
\end{table*}

\subsection{Data Collection}

We use a Chrysler Pacifica minivan mounted with a sensor suite of six cameras, two LiDARs, and one GNSS/IMU device to collect data in Silicon Valley. Five cameras cover a $360\degree$ area, while a mechanical spinning LiDAR (Pandar64, $200m$ range at $10\%$ reflectivity) and a forward-facing LiDAR (PandarGT, $300m$ range at $10\%$ reflectivity) enable much longer 3D object detection range to better support high-speed autonomous driving scenarios. See Figure \ref{fig:Sensor Setup for dataset acquisition} for the sensor layout, Table \ref{tab:The specifications of sensors} for detailed sensor specifications, and Figure \ref{fig:Overview of PandaSet} for point cloud samples.

A frame-based data structure is used to encapsulate point cloud and image data. One frame of the image refers to a single picture taken after the camera is exposed to light. One frame of the point cloud refers to a point cloud set obtained after the LiDAR completes a scan cycle. The mechanical spinning LiDAR sweeps in a $360\degree$ circle with $10Hz$ frequency, while the forward-facing LiDAR uses MEMS mirror-based scanning technology with $10Hz$ frequency. To achieve better data alignment between the LiDARs and cameras, we use a trigger board to trigger each camera to expose only when the mechanical spinning LiDAR scans across the center of a specific camera's FOV, ensuring the camera and the LiDAR capture the same objects at the same time.

The timestamp of each image is the exposure time, calculated by adding the exposure trigger time to the exposure duration. The exposure trigger time is estimated by the timestamp when the mechanical spinning LiDAR sweeps across the center of the camera's FOV. The exposure duration is estimated by test statistics. Since the cameras used are all automatic exposure-controlled, we use two different exposure duration parameters for daytime and nighttime to provide more accurate estimations. The timestamp of the point cloud frame is the time that the LiDAR takes to complete the scan cycle. Moreover, each point's timestamp is provided in the frame's specific information. We use PTP for time synchronization of the two LiDARs. The time source for the entire suite is GPS clock.

%\begin{figure}[!b]
%        \vspace{-10pt}
%        \begin{center}
%          \subfigure[]{\label{fig:projcted pc}\includegraphics[scale=0.261]{image/projected-pc.pdf}}
%          \subfigure[]{\label{fig:spliced pc}\includegraphics[scale=0.28]{image/spliced-pc.pdf}}
%        \end{center}
%        \vspace{-10pt}
%        \caption{Sample data from PandaSet. (a) Taken from a $360\degree$ LiDAR sweep. Camera images overlaid with LiDAR point cloud. The upper middle image features point cloud from the forward-facing LiDAR; all other images feature point cloud from the mechanical spinning LiDAR. (b) Point cloud alignment of sequential LiDAR scans}
        % \caption{Sample from PandaSet, taken from a 360-degree LiDAR sweep. Camera images overlaid with LiDAR point cloud. The upper middle image features point cloud from the forward-facing LiDAR; all other images feature point cloud from the mechanical spinning LiDAR.}
%        \label{fig:projected and spliced}
        %\vspace{-10pt}
%\end{figure}

\subsection{Sensor Calibration}

\begin{figure}[t]
        \begin{center}
                \subfigure[]{\label{fig:sample a projected image }\includegraphics[scale=0.9]{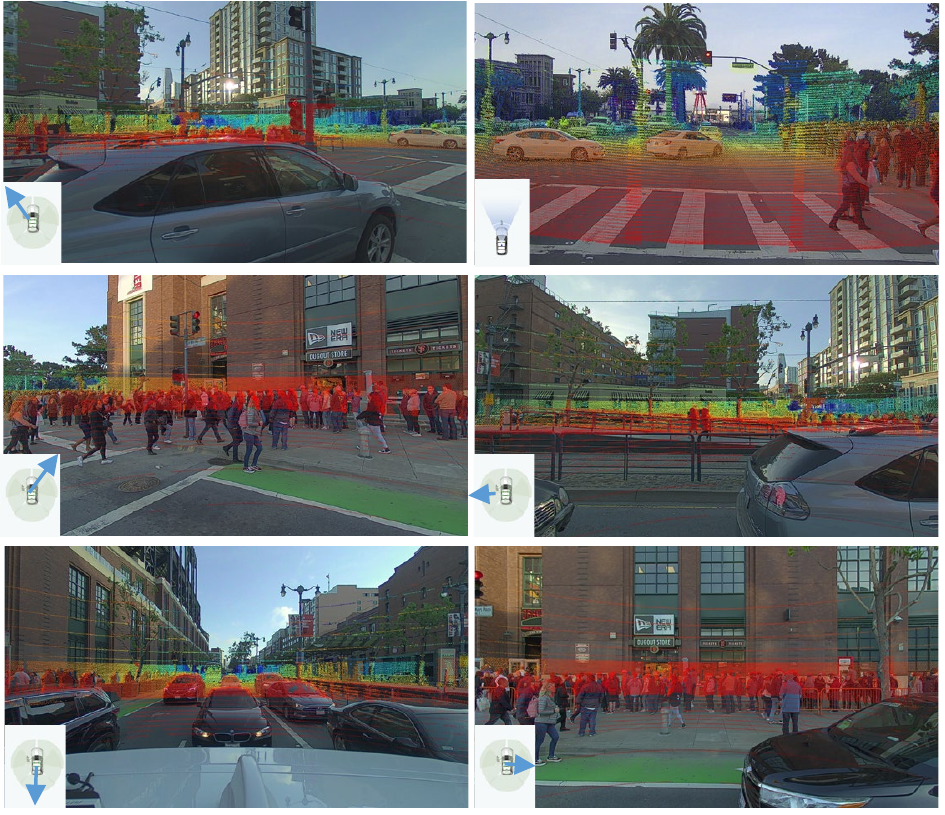}} \\
                % \subfigure[]{\label{fig:sample b projected point cloud}\includegraphics[scale=0.35]{image/2.pdf}} \\
                \subfigure[]{\label{fig:sample b projected point cloud}\includegraphics[scale=0.28]{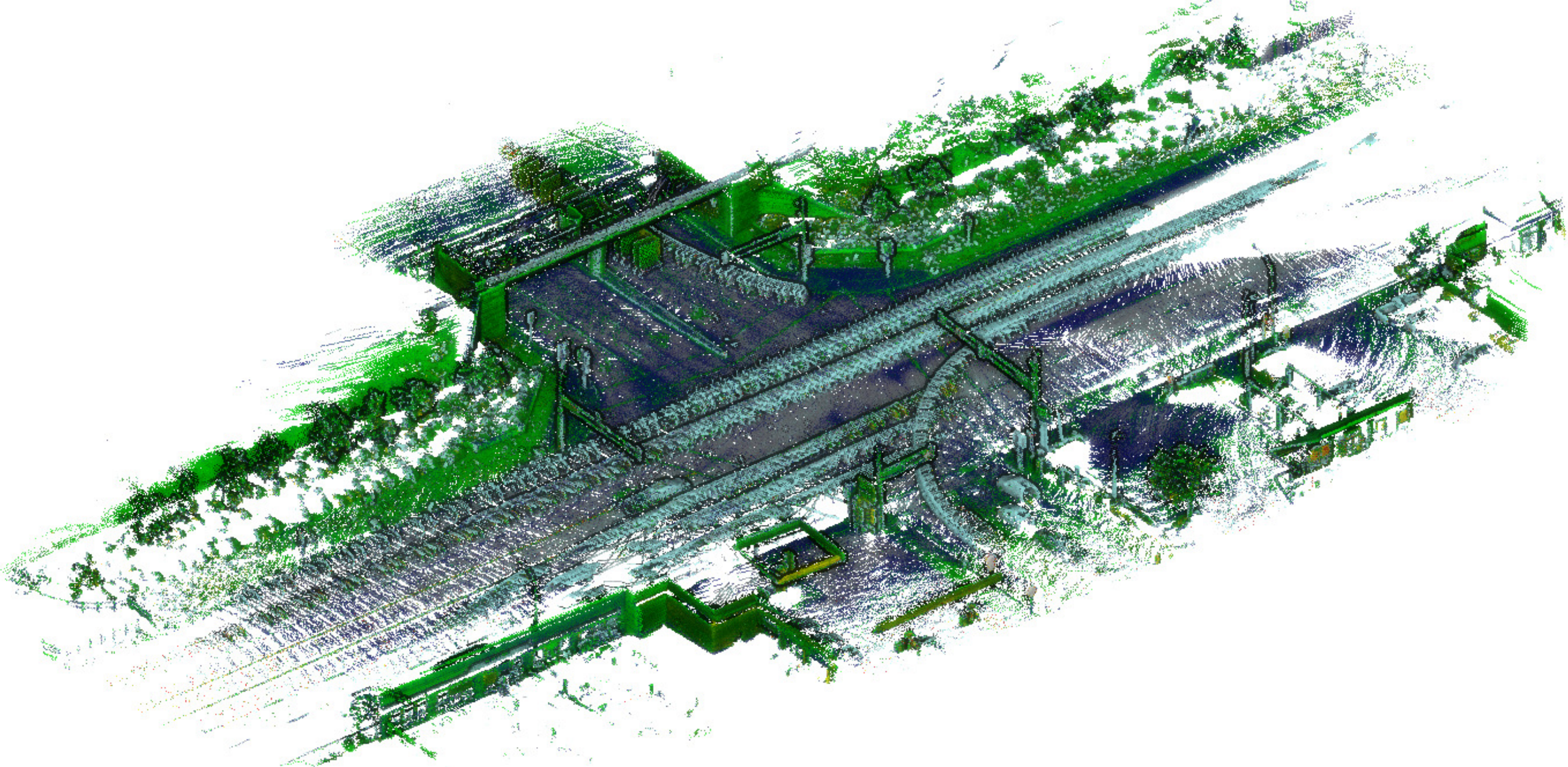}} \\
        \end{center}
        \vspace{-10pt}
        \caption{Sample data from PandaSet. (a): Taken from a $360\degree$ LiDAR sweep. Camera images overlaid with LiDAR point cloud. The top right image features point cloud from the forward-facing LiDAR; all other images feature point cloud from the mechanical spinning LiDAR. (b): Point cloud alignment of sequential LiDAR scans}
        \begin{center}
              %  \vspace{-5pt}
              %  \footnotesize{(grids in figure length 20m)}
        \end{center}
        \label{fig:projected point cloud sample}
\vspace{-20pt}
\end{figure}

\begin{figure}[t]
        %\vspace{-10pt}
        \begin{center}
          \includegraphics[scale=0.27]{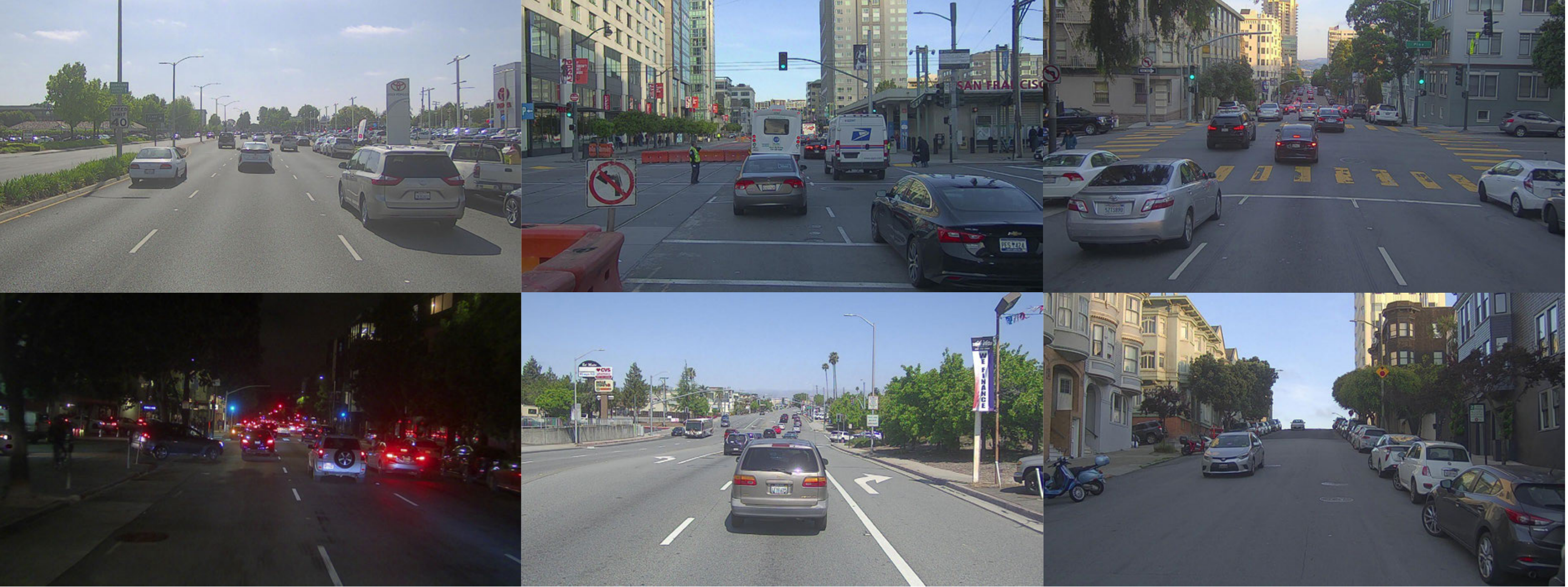}
        \end{center}
        \vspace{-5pt}
        \caption{Sample images from PandaSet capturing various lighting conditions and road environments.}
        % \caption{Sample from PandaSet, taken from a 360-degree LiDAR sweep. Camera images overlaid with LiDAR point cloud. The upper middle image features point cloud from the forward-facing LiDAR; all other images feature point cloud from the mechanical spinning LiDAR.}
        \label{fig:projected and spliced}
        % \vspace{-15pt}
\end{figure}

To ensure high quality with a multi-sensor dataset, it is important to calibrate the extrinsics and intrinsics of each sensor. The sensor calibration referred to in PandaSet includes intrinsic calibration of the cameras, extrinsic calibration of Pandar64-to-camera, extrinsic calibration of PandarGT-to-Pandar64, and extrinsic calibration of Pandar64-to-GNSS/IMU. Since all sensors are mounted tightly on our vehicle and the entire collection process was completed within two days, we assume the intrinsic and extrinsic parameters of the sensors remain unchanged. In other words, PandaSet has only one set of sensor intrinsic and extrinsic parameters. Moreover, to implement motion compensation, we estimate the vehicle's ego motion at each timestamp of the point cloud with linear interpolation of the vehicle's GNSS/IMU data, helping better align LiDAR scans and images, as well as consecutive LiDAR scans. Results are shown in Figure \ref{fig:projected point cloud sample}. Note that all point cloud data in PandaSet is based on a global coordinate system rather than an ego coordinate system. Each sequence has its own definition of the global coordinate system with the origin at the vehicle's start position.

\subsection{Scenes Selection}

From the dataset's $103$ scenes ($8$ seconds each), there are a total of $41,200$ wide-angle images, $8240$ telephoto images, $8240$ point cloud frames of the mechanical spinning LiDAR, and $8240$ point cloud frames of the forward-facing LiDAR. All scenes are carefully selected to cover different driving conditions including complex urban environments (e.g. dense traffic, pedestrians, construction), uncommon object classes (e.g. construction vehicles, motorized scooters), a diversity of roads and terrain (e.g. sharp turns, hills), and different lighting conditions throughout the day and at night. The diversity and complexity of the scenes help capture the complex, varied scenarios of real-world driving. Raw data is collected from two routes in Silicon Valley: (1) San Francisco, and (2) El Camino Real from Palo Alto to San Mateo.

\subsection{Data Annotation}

PandaSet provides high-quality ground truth annotations of its sensor data, including 3D bounding box labels for all $103$ scenes and point cloud semantic segmentation annotations for $76$ scenes for both mechanical spinning and forward-facing LiDARs. The annotation frequency remains $10Hz$. For 3D object detection, we annotate 3D bounding boxes for $28$ object classes (e.g. cars, buses, motorcycles, traffic cones) with a rich set of class attributes related to activity, visibility, location, and pose. All cuboids contain at least $5$ LiDAR points, except the ones for which we can accurately predict the size and location for occluded or distant paths (this exception only applies if there is, at minimum, one frame where the same object has at least $5$ LiDAR points.) For the task of point cloud semantic segmentation, we annotate points with $37$ different semantic labels (e.g. car exhaust, lane markings, drivable surfaces). Based on pixel-level sensor fusion technology, we combine multiple LiDAR and camera inputs into one point cloud, enabling the highest precision and quality annotations.

%\begin{figure}[b]
%        \begin{center}
%        \includegraphics[scale = 0.26]{image/Num-type.pdf}
%        \end{center}
%        \vspace{-10pt}
%        \caption{Average numbers of different types in each frame}
%        \label{fig:Num of types vs KITTI}
%        \vspace{-10pt}
%\end{figure}

%\begin{figure}[b]
%        \begin{center}
%                \includegraphics[scale=0.25]{image/Num-distance.pdf}        
%        \end{center}
%        \vspace{-15pt}
%        \caption{Statistics of object numbers in each frame with distance}
%        \label{fig:Nums with distance VS KITTI}
%        \vspace{-10pt}
%\end{figure}

\begin{table}[t]
        \caption{Comparison with Waymo Open dataset of labeled traffic participants (instances per frame)}
        \renewcommand\arraystretch{1.5}
        \vspace{-15pt}
        \scriptsize
        \label{tab:VS Waymo}
        \begin{center}
        \begin{tabular}{|ccccc|}
        \multicolumn{5}{c}{($\ssymbol{2}$:Distance \textless $75m$)}\\[0pt]
        \hline
        & \textbf{Car} & \textbf{Pedestrian} & \textbf{Cyclist} & \textbf{Sign} \\
        \hline
        \hline
        %Waymo & 6.1M & 2.8M & 67k & 3.2M  \\
        Waymo & $26.5$ & $12.2$ & $0.3$ & $\mathbf{13.9}$  \\
        \hline
        %\textbf{PandaSet} & \textbf{1.1M} & \textbf{249k} & \textbf{32k} & \textbf{25k}  \\
        PandaSet$\ssymbol{2}$ & $67.0$ & $20.3$ & $2.8$ & $1.8$ \\
        \hline
        %PandaSet\textless75m & 552k & 167k & 23k & 16k  \\
        PandaSet & $\mathbf{133.5}$ & $\mathbf{30.2}$ & $\mathbf{3.9}$ & $3.0$  \\
        \hline
         
        \end{tabular}
        \end{center}
        \vspace{-15pt}

\end{table}

\begin{figure}[b]
        \begin{center}
        \includegraphics[width=7.5cm]{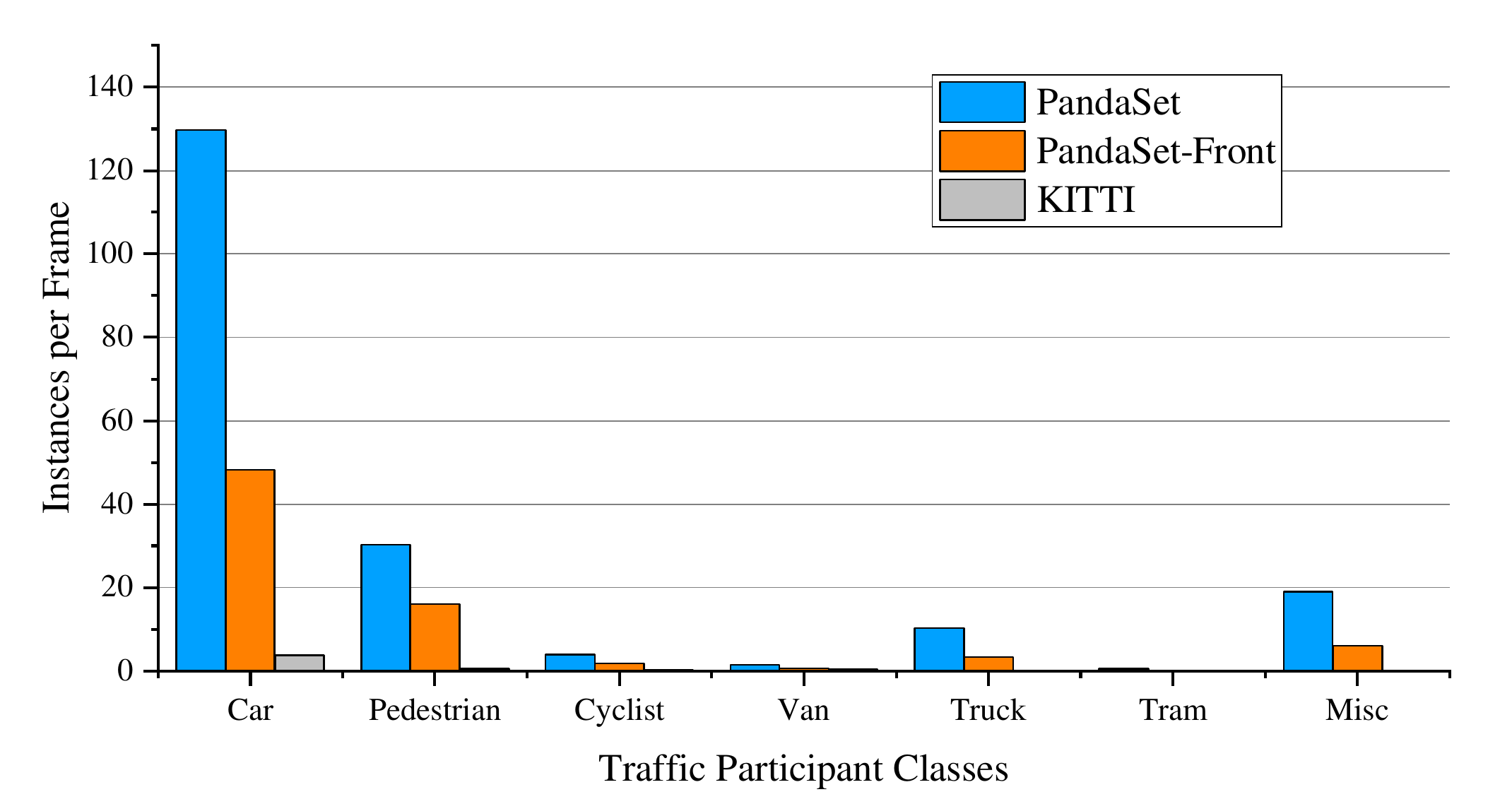}
        \end{center}
        \vspace{-10pt}
        %\caption{Average numbers of different types in each frame}
        \caption{Instances per frame of $7$ traffic participant classes. To stay consistent with KITTI, which only labels objects within its front-facing camera's FOV, PandaSet-Front only includes instances of objects within that same FOV.}
        \label{fig:Num of types vs KITTI}
        % \vspace{-10pt}
\end{figure}

\begin{figure}[t]
        \begin{center}
                \includegraphics[width=7.5cm]{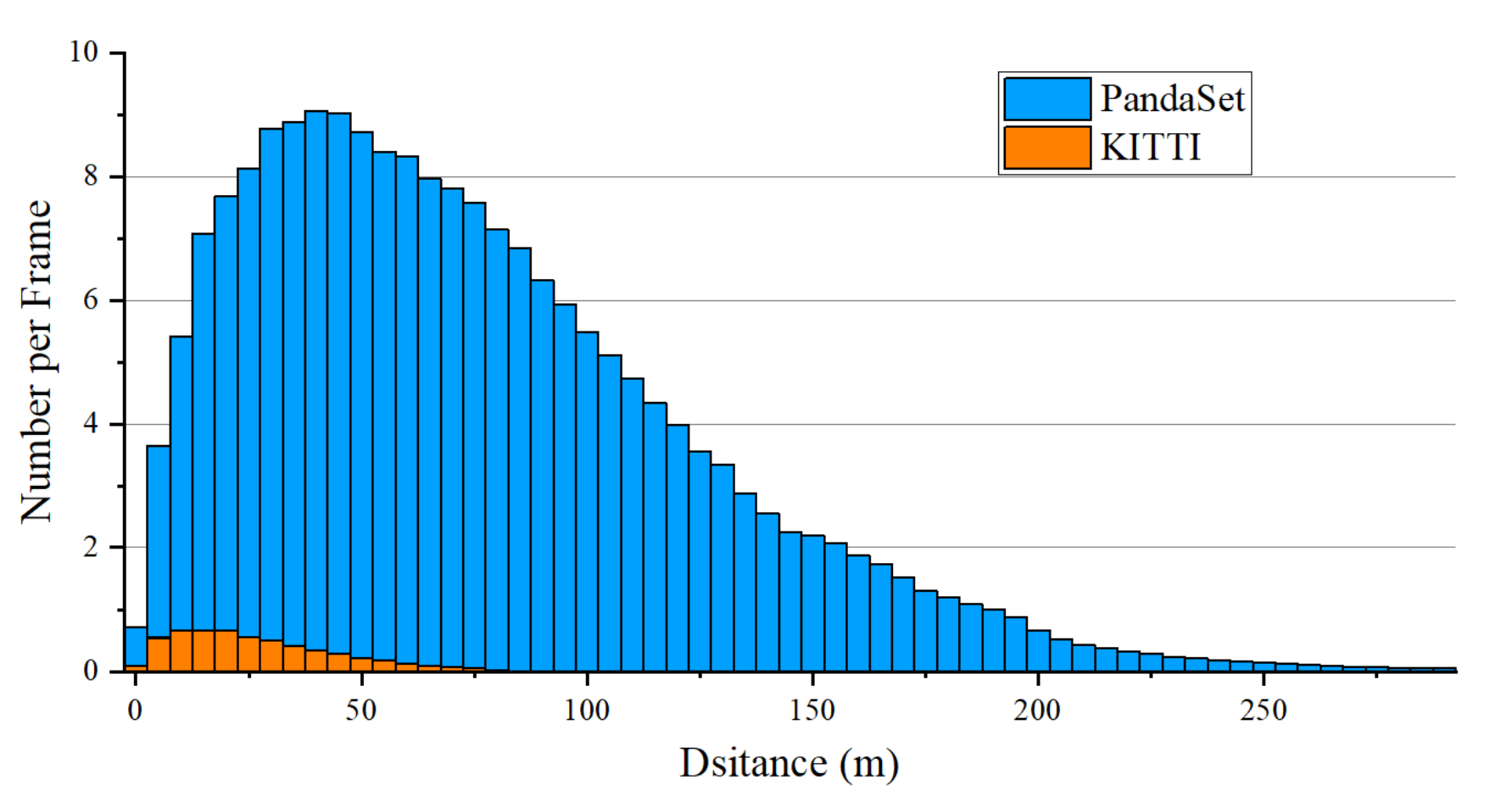}  
        \end{center}
        \vspace{-15pt}
        %\caption{Statistics of object numbers in each frame with distance}
        \caption{Instances of all traffic participants per frame, distributed over distance.}
        \label{fig:Nums with distance VS KITTI}
        \vspace{-15pt}
\end{figure}

\begin{figure}[!t]
        \begin{center}
          \subfigure[Total instances of each object class in PandaSet. Ped = Pedestrian; Ped$^{\ast}$ = Pedestrian with object; M Truck = Medium-Sized Truck; RC = Rolling Container.
          ]{\label{fig:number of objects in each category}\includegraphics[width=7.5cm]{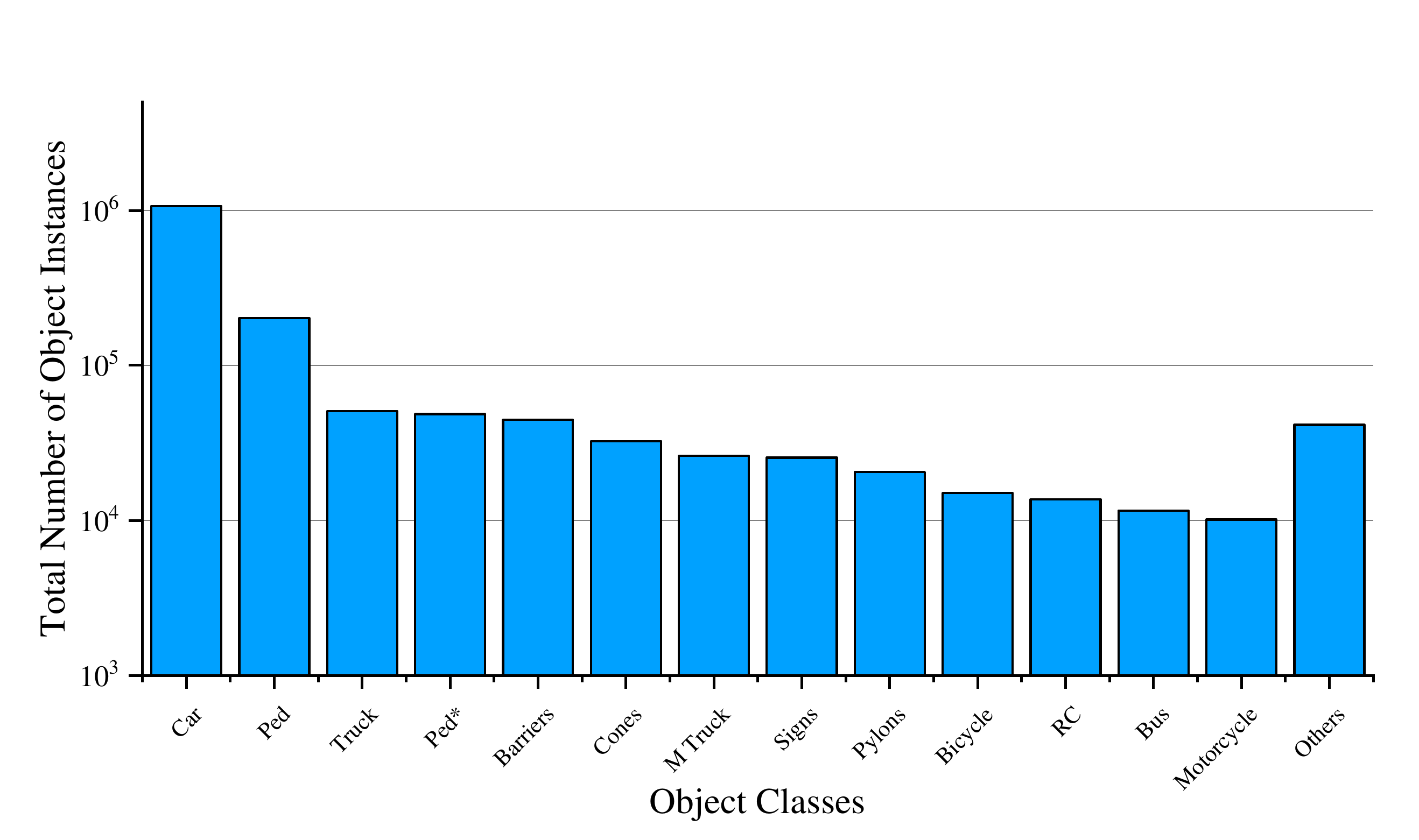}}
          \subfigure[Total number of LiDAR points for each semantic segmentation class in PandaSet. Other S-O = Other Static Object; M Truck = Medium-Sized Truck; Other R-M = Other Road Marking; LLM = Lane Line Marking.]{\label{fig:point numbers for each category}\includegraphics[width=7.5cm]{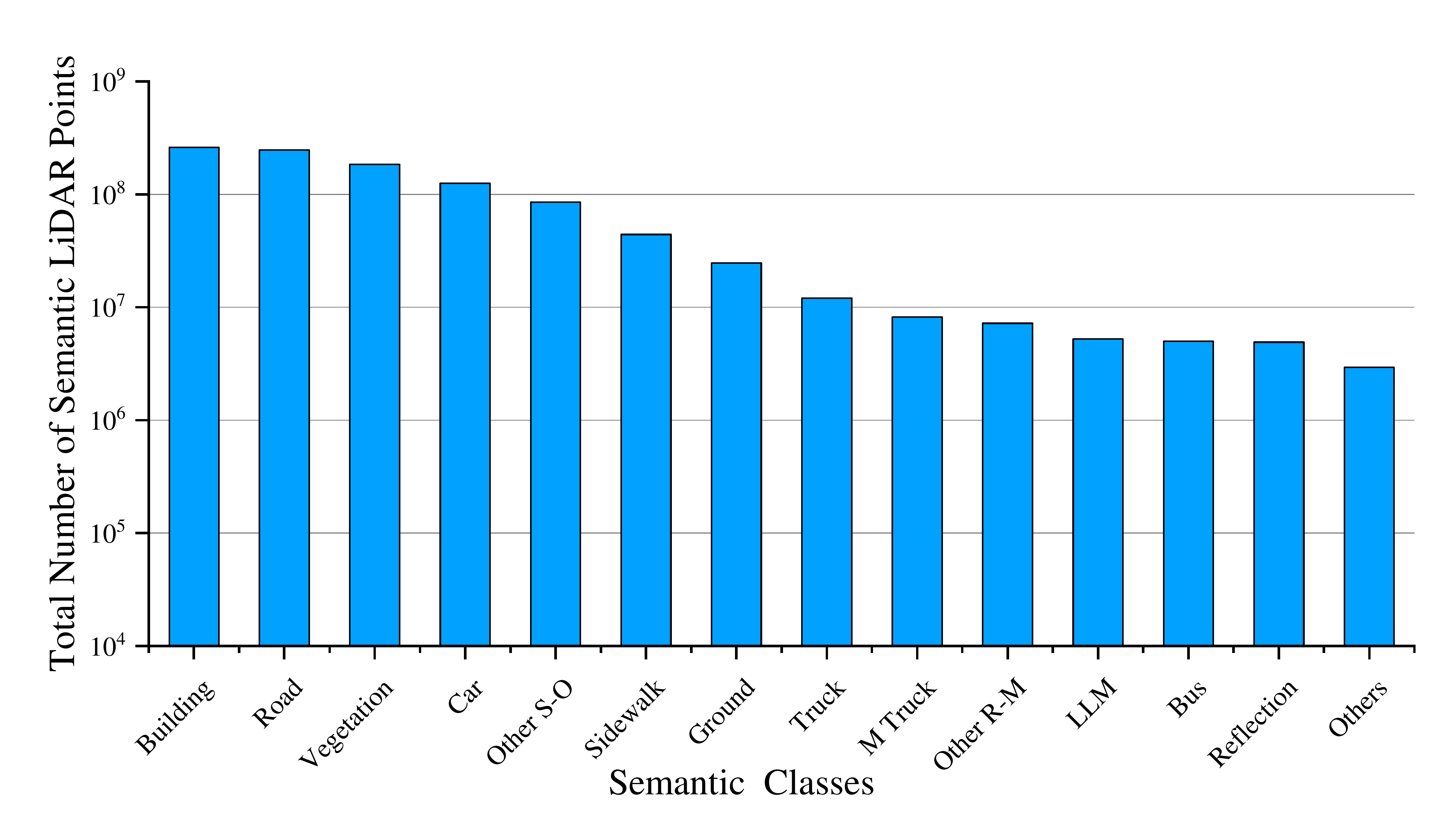}} \\
          \subfigure[Proportion of attribute annotations for Car (left) and Pedestrian (right). Left: P = Parked; S = Stopped; M = Moving. Right: St = Standing; W = Walking; Si = Sitting; L = Lying.]{\label{fig:Car}\includegraphics[width=7.5cm]{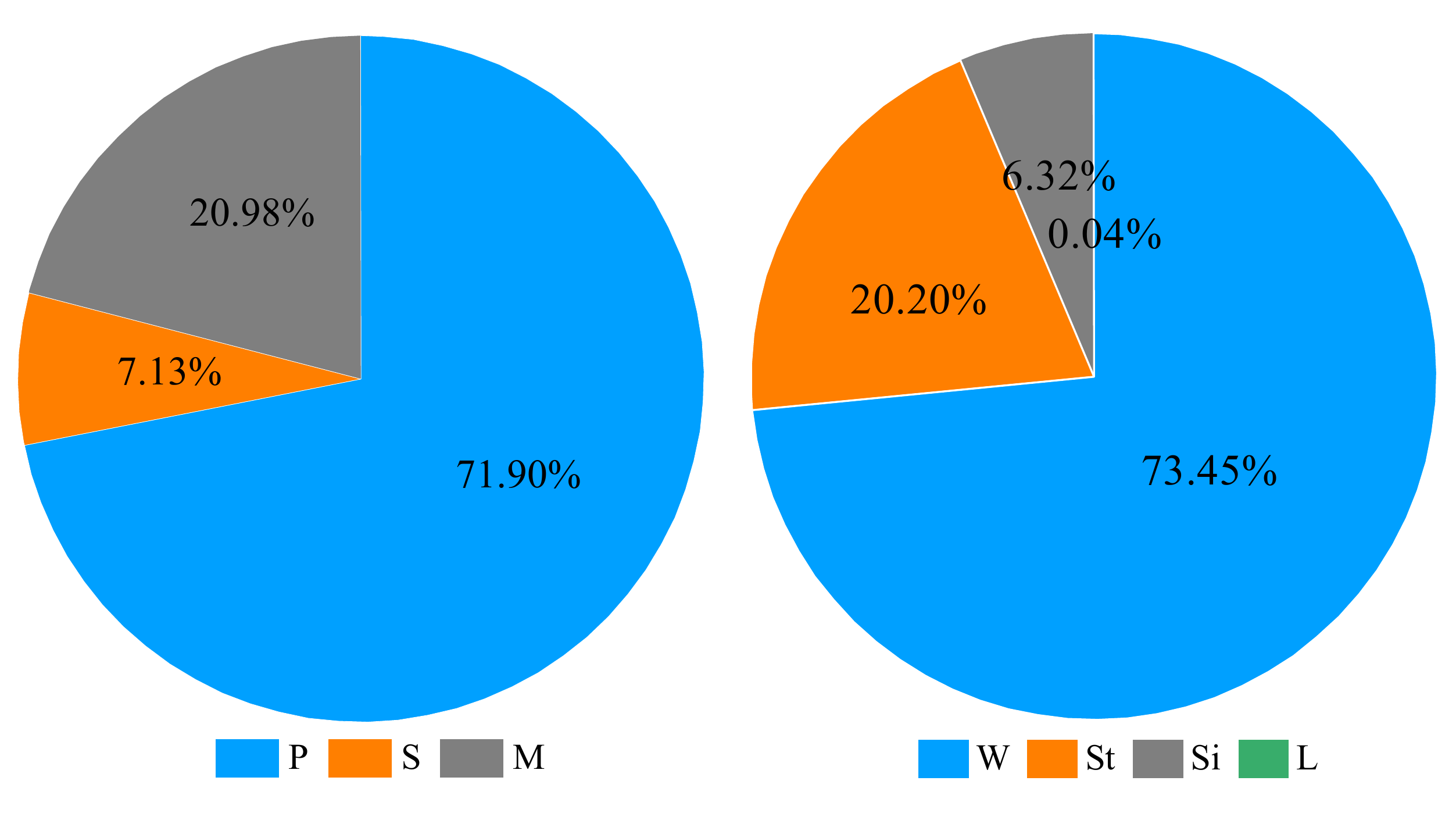}}
        % \subfigure[tODO]{\label{fig:Car}\includegraphics[width=7.5cm]{image/Num-type.pdf}}
          %\subfigure[Proportion of attribute annotations for Pedestrian]{\label{fig:Pedstrian}\includegraphics[scale=0.23]{image/Pro-of-Ped.pdf}}
        \end{center}
        \vspace{-10pt}
        \caption{Data statistics of PandaSet}
        \label{fig:Data statistics of PandaSet}
\vspace{-15pt}
\end{figure}

\subsection{Dataset Statistics}

By combining the strengths of a complete, high-precision sensor kit, particularly the long-range mechanical spinning and forward-facing LiDARs, we can annotate objects up to $300$ meters, significantly further than most other datasets. See Figure \ref{fig:Nums with distance VS KITTI} for the average distance distribution of objects per frame, compared with KITTI. Moreover, we provide object annotations in full $360\degree$ view, as opposed to only in frontal view. These features enable PandaSet to achieve higher label density. See Figure \ref{fig:Num of types vs KITTI} and Table \ref{tab:VS Waymo} for a comparison with KITTI and Waymo Open dataset of main traffic participant annotations. Furthermore, PandaSet has the largest number of label categories and the most elaborate label taxonomy among current open-source datasets, as shown in Table \ref{tab:Current autonomous driving datasets}. Rare object classes such as motorized scooters, rolling containers, animals (e.g. birds), smoke, and car exhaust can provide a useful resource for researchers to address the long tail in real-world driving scenarios, which is a major challenge for the safe deployment of autonomous vehicles. See Figure \ref{fig:Data statistics of PandaSet} for the statistics of the annotated categories in both 3D object detection and point cloud semantic segmentation.

\begin{table}[t]
        \caption{Object Detection Difficulty Rating}
        \renewcommand\arraystretch{1.5}
        \vspace{-10pt}
        \label{tab:Level difficulty}
        \begin{center}
        \begin{tabular}{|p{1.2cm}<{\centering}|p{1.2cm}<{\centering}|p{1.2cm}<{\centering}p{1.2cm}<{\centering}|}
                % \multicolumn{6}{c}{(Note: The distances expressed in the table signify a range ($50m$ is $0$ - $50m$; $70m$ is $50$ - $70m$, and so on).)}\\[0pt]
                % \multicolumn{6}{c}{\footnotesize TABLE V:  }\\[1pt]
        \hline
        % \textbf{LiDAR} & \textbf{Category} & \textbf{LEVEL\_1} & \textbf{LEVEL\_2}  \\
        \textbf{LiDAR} & \textbf{Category} & \textbf{LEVEL1} & \textbf{LEVEL2}  \\
        \hline
        \hline
        \multirow{3}{*}[-0.5ex]{Pandar64} %&Car & BEV@0.7 & - & 80.35 & 79.48 & 69.52 & 60.91 \\ 
        %\cline{2-8}
        & Car & $91.8\%$ & $8.2\%$ \\
        %\cline{2-8}
          % & Pedestrian & BEV@0.5 & - & 38.65 & 41.08 & 36.09 & 31.13 \\
        %\cline{2-8}
        & Pedestrian & $86.7\%$ & $13.3\%$ \\
        %\cline{2-8}
         % & Cyclist & BEV@0.5 & - & 37.65 & 30.07 & 17.97 & 14.16  \\
        %\cline{2-8}
        % & Sign & 92.30 & 7.70 \\
        & Cyclist & $86.8\%$ & $13.2\%$ \\
        % & Misc & 80.43 & 19.57 \\
        \hline
        \multirow{3}{*}[-0.5ex]{PandarGT} & Car & $74.1\%$ & $25.9\%$ \\
        & Pedestrian & $69.9\%$ & $30.1\%$ \\
        & Cyclist & $79.2\%$ & $29.8\%$ \\
            %& Pedestrian & BEV@0.5 & 58.02 & 57.21 & - & - &- \\
        %\cline{2-8}
        % & Sign & 78.8 & 21.20 \\
        %\cline{2-8}
            %& Cyclist & BEV@0.5 & 55.80 & 52.93 & - & - &- \\
        %\cline{2-8}
        % & Misc & 74.14 & 25.86 \\
        \hline
\end{tabular}
\end{center}
\vspace{-20pt}
\end{table}

%%%%%%%%%%%%%%%%%%%%%%%%%%%%%%%%%%%%%%%%%%%%%%%%%%%%%%%%%%%%%%%%%%%%%%%%%%%%%%%%%%%%%%%%%%%%%%%%

\section{BASELINE EXPERIMENTS}
\label{sec:BASELINE EXPERIMENTS}

We establish baselines on our dataset with methods for LiDAR-only 3D object detection, LiDAR-camera fusion 3D object detection, and LiDAR point cloud segmentation. We select the first $50$ frames of data from each sequence as training data and the remaining $30$ frames as test data to make a tradeoff between robustness and uniformity of the evaluation method. Since we have $103$ sequences for 3D object detection and $76$ sequences for LiDAR point cloud segmentation, there are $5150$ training samples and $3090$ test samples for 3D object detection, $3800$ training samples, and $2280$ test samples for LiDAR point cloud segmentation.

\subsection{LiDAR-only 3D object detection}
\label{subsec:LiDAR-only 3D object detection}

To establish the baseline for LiDAR-only 3D object detection, we retrained PV-RCNN\cite{shi2020pv}, the top-performing network on both KITTI dataset and Waymo Open dataset. We use the publicly released code\footnote{https://github.com/open-mmlab/OpenPCDet} for PV-RCNN and keep it for only $3$ object classes (cars, pedestrians, and cyclists) to align with KITTI 3D object detection evaluation benchmark. Since the coverage of the mechanical spinning LiDAR and the forward-facing LiDAR is different, we retrained two different models with slight differences in network configuration. The detection range along the x-axis is set to $[-70.4m, 70.4m]$ for the mechanical spinning LiDAR and $[0m, 211.2m]$ for the forward-facing LiDAR. Other configuration parameters are shared in both models. The detection range along the y-axis is set to $[-51.2m, 51.2m]$ and $[-2m, 4m]$ along the z-axis. The voxel size is set to $(0.1m, 0.1m, 0.15m)$. Both LiDARs' point cloud frame data is based on the ego vehicle frame, whose x-axis is positive in the forward direction, y-axis is positive to the left, and z-axis is positive in the upward direction. In the 3D proposal generation module, we define anchor sizes $(l, w, h)$ as $(3.9m, 1.6m, 1.56m)$ and $(5.02m, 2.0m, 1.82m)$ for cars, $(0.8m, 0.6m, 1.73m)$ for pedestrians, and $(1.76m, 0.6m, 1.73m)$ for cyclists. All object classes have anchors oriented to $0$ and $\pi/2$ radians. In the keypoints sampling module, we continue to use the Furthest-Point-Sampling (FPS) algorithm to sample $8192$ points, to be consistent with the original paper.

\begin{table}[t]
        %\caption{AP performance of Pandar64 and PandarGT}
        \caption{Baseline 3D AP for LiDAR-only 3D object detection}
        %\caption{AP PERFORMANCE OF PANDAR64 AND PANDARGT}
        \renewcommand\arraystretch{1.5}
        \vspace{-5pt}
        \label{tab:AP of Pandar64 and PandarGT}
        % \begin{center}
        {\centering
        \begin{tabular}{|p{1.1cm}<{\centering}|p{1.1cm}<{\centering}|p{0.45cm}<{\centering}p{0.45cm}<{\centering}p{0.45cm}<{\centering}p{0.5cm}<{\centering}p{0.7cm}<{\centering}|}
                % \multicolumn{7}{c}{(Note: The distances expressed in the table signify a range} \\
                % \multicolumn{7}{c}{($50m$ is $0$ - $50m$; $70m$ is $50$ - $70m$, and so on).)}\\[0pt]
                % \multicolumn{6}{c}{\footnotesize TABLE V:  }\\[1pt]
        \hline
        \textbf{LiDAR} & \textbf{Category} & $\mathbf{50m}$ & $\mathbf{70m}$ & $\mathbf{100m}$ & $\mathbf{150m}$ & $\mathbf{200m}$ \\
        \hline
        \hline
        \multirow{3}{*}[-0.5ex]{Pandar64} %&Car & BEV@0.7 & 89.14 & 80.84 & - & - & -   \\ 
            & Car  & $79.56$ & $78.36$ & - & - &- \\
            %& Pedestrian & BEV@0.5 & 58.02 & 57.21 & - & - &- \\
        %\cline{2-8}
            & Pedestrian & $55.37$ & $55.70$ & - & - &- \\
        %\cline{2-8}
            %& Cyclist & BEV@0.5 & 55.80 & 52.93 & - & - &- \\
        %\cline{2-8}
            & Cyclist & $53.33$ & $52.41$ & - & - &-  \\
        \hline
        \multirow{3}{*}[-0.5ex]{PandarGT} %&Car & BEV@0.7 & - & 80.35 & 79.48 & 69.52 & 60.91 \\ 
        %\cline{2-8}
          & Car  & - & $73.59$ & $69.34$ & $57.92$ & $49.57$ \\
        %\cline{2-8}
          % & Pedestrian & BEV@0.5 & - & 38.65 & 41.08 & 36.09 & 31.13 \\
        %\cline{2-8}
          & Pedestrian  & - & $32.07$ & $33.99$ & $31.00$ & $27.68$ \\
        %\cline{2-8}
         % & Cyclist & BEV@0.5 & - & 37.65 & 30.07 & 17.97 & 14.16  \\
        %\cline{2-8}
          & Cyclist  & - & $34.28$ & $28.50$ & $17.26$ & $13.99$  \\
        \hline
\end{tabular}
% \end{center}
\par}\medskip
\vspace{-5pt}
\end{table}

The results of the test set are evaluated by average precision (AP), the commonly used evaluation benchmark with $11$ recall positions\cite{geiger2012we}. We use $0.7$ IoU threshold for cars and $0.5$ IoU threshold for pedestrians and cyclists in 3D test. See Table \ref{tab:AP of Pandar64 and PandarGT} for detailed results. The distances expressed in the table signify a range ($50m$ is $0$--$50m$; $70m$ is $50$--$70m$, and so on). The decline in detection performance of pedestrians and cyclists with the forward-facing LiDAR's model is likely due to an increased level of detection difficulty, as shown in Table \ref{tab:Level difficulty}. The difficulty ratings, LEVEL\_1 and LEVEL\_2, use Waymo Open dataset's difficulty definitions for the single frame 3D object detection task\cite{sun2020scalability}. Examples with ${\leq}5$ LiDAR points are designated as the more challenging LEVEL\_2.

%\begin{table}[t]
%        \caption{}
%        \renewcommand\arraystretch{1.5}
%        \vspace{-10pt}
%        \scriptsizetab:7 & 79.56 & 78.36  \\
%        \hline
%        Pedestrian & BEV@0.5 & 58.02 & 57.21 \\
%        \hline
%        Pedestrian & 3D@0.5 & 55.37 & 55.70  \\
%        \hline
%        Cyclist & BEV@0.5 & 55.80 & 52.93  \\
%        \hline
%        Cyclist & 3D@0.5 & 53.33 & 52.41   \\
%        \hline
%        \end{tabular}
%        \end{center}
%\end{table}

\begin{table}[t]
        %\caption{PointPainting on PandaSet vs kitti, 3D AP}
        \caption{Baseline 3D AP for LiDAR-camera fusion 3D object detection}
        \renewcommand\arraystretch{1.5}
        \vspace{-5pt}
        \scriptsize
        \label{tab:PointPainting on PandaSet vs kitti, 3D AP}
        % \begin{center}
        {\centering
        \begin{tabular}{|cccc|}
        %\multicolumn{4}{c}{\footnotesize TABLE IV: }\\[0pt]
        \hline
        \textbf{Dataset} & \textbf{Car} & \textbf{Pedestrian} & \textbf{Cyclist}  \\
        \hline
        \hline
        %-- & -- & -- & --  \\
        %\hline
        %KITTI & 77.74 & \textbf{61.67} & \textbf{71.62}  \\
        KITTI & $77.74$ & $61.67$ & $71.62$  \\
        \hline
        %\textbf{PandaSet} & \textbf{80.36} & 60.00 & 58.97 \\
        PandaSet & $80.36$ & $60.00$ & $58.97$ \\
        \hline
        \end{tabular}
        % \end{center}
        \par}\medskip
        \vspace{-20pt}
\end{table}

\subsection{LiDAR-camera fusion 3D object detection}

To establish the baseline for LiDAR-camera fusion 3D object detection, we re-implement PointPainting\cite{vora2020pointpainting}, an effective sequential architecture to fuse point clouds with semantic information from images. We use DeepLabv3+\cite{chen2018encoder} to output per-pixel class scores of the image and choose PointRCNN\cite{shi2019pointrcnn}\footnote{https://github.com/sshaoshuai/PointRCNN} as the 3D object detection network because of the strong performance shown in the original paper. Most configuration parameters remain unchanged from how they are presented in the paper, with the exception that we train a 19-classes-output DeepLabv3+ with the pretrained model for Cityscapes\cite{cordts2016cityscapes} supported by the publicly released code\footnote{https://github.com/NVIDIA/semantic-segmentation}. After image semantic segmentation, we merge the output from $19$ object classes into $4$ object classes (cars, pedestrians, cyclists, and background). Only the object examples in the field of view of the forward-facing, long-focus camera are involved in training and inference. We use the $50m$ range clipped point cloud from the mechanical spinning LiDAR for the LiDAR data input to align with the KITTI dataset. The same AP evaluation benchmark and IoU threshold are used for LiDAR-camera fusion 3D object detection as described in Section \ref{subsec:LiDAR-only 3D object detection}. See Table \ref{tab:PointPainting on PandaSet vs kitti, 3D AP} for detailed results. 

\subsection{LiDAR point cloud segmentation}

Since PandaSet also provides ground truth labels for LiDAR point cloud segmentation, we establish its baseline using RangeNet53, using only the network, without the post-processing\cite{milioto2019rangenet++}. The publicly released code\footnote{https://github.com/PRBonn/lidar-bonnetal} is used. With the whole network unchanged, we merge the $37$ final classes in the original output into $14$ primary classes for autonomous driving. In the inference, the class with the highest score represents the class ouput of each point. The commonly applied IoU matrix\cite{everingham2015pascal} is used as the evaluation matrix. See Table \ref{tab:RangeNet53 on PandaSet} for the detailed results.

\begin{table}[t]
        \caption{Baseline IoU metrics for LiDAR point cloud segmentation}
        %\caption{RANGENET53 ON PANDASET}
        \renewcommand\arraystretch{1.5}
        \vspace{-10pt}
        \scriptsize
        \label{tab:RangeNet53 on PandaSet}
        \begin{center}
        \begin{tabular}{|lclc|}
        %\multicolumn{4}{c}{\footnotesize TABLE IV: }\\[0pt]
        \hline
        \textbf{Type} & \textbf{IoU} & \textbf{Type} & \textbf{IoU} \\
        \hline
        \hline
        Car & $0.875$ & Pedestrian & $0.374$ \\
        \hline
        Motorcycle & $0.428$ & Pickup Truck & $0.701$ \\
        \hline
        Bus & $0.699$ & Other Vehicle & $0.388$ \\
        \hline
        Train & $0.807$ & Animals-Bird & $0.276$ \\
        \hline
        Road Barriers & $0.718$ & Signs & $0.325$ \\
        \hline
        Building & $0.929$ & Ground & $0.587$ \\
        \hline
        Road & $0.862$ & Lane Line Marking & $0.426$ \\
        %-- & -- & -- & --  \\
        %\hline
        \hline
        \end{tabular}
        \vspace{-15pt}
        \end{center}
\end{table}

\section{CONCLUSION}
\label{sec:CONCLUSION}

In this paper, we introduce PandaSet, the world's first open-source dataset to include both mechanical spinning and forward-facing LiDARs and to be released free-of-charge for both research and commercial use. We present details on data collection and annotation. At a time when the barriers to data collection are still high, PandaSet was released in the hopes of helping the broader research and developer community accelerate the safe deployment of autonomous vehicles. In the future, we plan to design evaluation metrics and build a public leaderboard to track the research progress in 3D detection and point cloud segmentation, and add map information to the dataset.

\addtolength{\textheight}{-8cm}   % This command serves to balance the column lengths
                                  % on the last page of the document manually. It shortens
                                  % the textheight of the last page by a suitable amount.
                                  % This command does not take effect until the next page
                                  % so it should come on the page before the last. Make
                                  % sure that you do not shorten the textheight too much.

%%%%%%%%%%%%%%%%%%%%%%%%%%%%%%%%%%%%%%%%%%%%%%%%%%%%%%%%%%%%%%%%%%%%%%%%%%%%%%%%

%%%%%%%%%%%%%%%%%%%%%%%%%%%%%%%%%%%%%%%%%%%%%%%%%%%%%%%%%%%%%%%%%%%%%%%%%%%%%%%%

%%%%%%%%%%%%%%%%%%%%%%%%%%%%%%%%%%%%%%%%%%%%%%%%%%%%%%%%%%%%%%%%%%%%%%%%%%%%%%%%
\section*{ACKNOWLEDGMENT}

The dataset analysis and baseline experiments presented in this paper were supported by the National Key Research and Development Program of China (2018YFB0105000). From Hesai, we thank Ziwei Pi and Congbo Shi for hardware design and implementation. The PandaSet dataset was annotated by Scale AI, supported by Dave Morse, Kathleen Cui, and Shivaal Roy.

\bibliographystyle{IEEEtran}
\bibliography{root}

\end{document}